\documentclass{article}

\usepackage[square,numbers]{natbib}  % for number citations

\usepackage[preprint]{neurips_2024}

\usepackage[utf8]{inputenc} % allow utf-8 input
\usepackage[T1]{fontenc}    % use 8-bit T1 fonts
\usepackage{hyperref}       % hyperlinks
\usepackage{url}            % simple URL typesetting
\usepackage{booktabs}       % professional-quality tables
\usepackage{amsfonts}       % blackboard math symbols
\usepackage{nicefrac}       % compact symbols for 1/2, etc.
\usepackage{microtype}      % microtypography
\usepackage{xcolor}         % colors

\usepackage{amsmath}
\usepackage{graphicx}
\usepackage{svg}
\usepackage[linesnumbered,ruled]{algorithm2e}
\DontPrintSemicolon % to let us print blank (numbered) lines in algorithm2e
\usepackage{multirow}  % for multi-row cells in a table
\usepackage{siunitx}  % for well-formatted scientific notation
\usepackage{mathtools}  % for \coloneq symbol

\newcommand{\Fig}[1]{Fig.~\ref{fig:#1}}
\newcommand{\Tab}[1]{Table~\ref{tab:#1}}
\newcommand{\Sec}[1]{Section~\ref{sec:#1}}
\newcommand{\App}[1]{Appendix~\ref{sec:#1}}
\newcommand{\Alg}[1]{Algorithm~\ref{alg:#1}}
\newcommand{\code}[1]{\texttt{#1}}

\title{LTAU-FF: \underline{L}oss \underline{T}rajectory \underline{A}nalysis for \underline{U}ncertainty in Atomistic \underline{F}orce \underline{F}ields}

\author{%
    Joshua A. Vita \\
    Materials Science Division \\
    Lawrence Livermore National Laboratory \\
    Livermore, California 94550, USA \\
    \And
    Amit Samanta \\
    Physics Division \\
    Lawrence Livermore National Laboratory \\
    Livermore, California 94550, USA \\
    \And
    Fei Zhou \\
    Physics Division \\
    Lawrence Livermore National Laboratory \\
    Livermore, California 94550, USA \\
    \And
    Vincenzo Lordi \\
    Materials Science Division \\
    Lawrence Livermore National Laboratory \\
    Livermore, California 94550, USA \\
}

\begin{document}

\maketitle

\begin{abstract}
Model ensembles are effective tools for estimating prediction uncertainty in deep learning atomistic force fields.
However, their widespread adoption is hindered by high computational costs and overconfident error estimates.
In this work, we address these challenges by leveraging distributions of per-sample errors obtained during training and employing a distance-based similarity search in the model latent space.
Our method, which we call \code{LTAU}, efficiently estimates the full probability distribution function (PDF) of errors for any test point using the logged training errors, achieving speeds that are 2--3 orders of magnitudes faster than typical ensemble methods and allowing it to be used for tasks where training or evaluating multiple models would be infeasible.
We apply \code{LTAU} towards estimating parametric uncertainty in atomistic force fields (\code{LTAU-FF}), demonstrating that its improved ensemble diversity
produces well-calibrated confidence intervals and predicts errors that correlate strongly with the true errors for data near the training domain.
Furthermore, we show that the errors predicted by \code{LTAU-FF} can be used in practical applications for detecting out-of-domain data, tuning model performance, and predicting failure during simulations.
We believe that \code{LTAU} will be a valuable tool for uncertainty quantification (UQ) in atomistic force fields and is a promising method that should be further explored in other domains of machine learning.

\end{abstract}

\section{Introduction} \label{sec:intro}

In computational chemistry and materials science, deep learning force fields have become an established tool for accelerating quantum mechanical calculations \cite{Deringer2019,Mueller2020,Mishin2021}.
The goal of these force fields is to perform the regression task of reproducing the potential energy surface of systems containing hundreds to millions or billions of atoms of one or more elemental types in a computationally efficient way \cite{NguyenCong2021,Musaelian2023b}.
This is done by learning to predict the energies and atomic forces from a high-accuracy ground truth, such as density functional theory (DFT).
A number of different approaches to this problem have been developed over the years, including the use of Gaussian processes \cite{Bartok2010,Christensen2020}, symbolic regression \cite{Hernandez2019}, feed-forward neural networks \cite{Behler2007,Manzhos2020}, and message-passing neural networks \cite{Batzner2022,Batatia2022a}, among others.
% However, with the transition away from more interpretable, classical physics-based force fields \cite{Jones1924,Baskes1992,vanDuin2001}, questions surrounding the necessary complexity of model architectures, and the domains over which they are valid, have become prevalent.
% Hence, developing UQ methods accounting for  different sources of epistemic uncertainty, therefore helping to isolate the effects of model uncertainty,
% % identifying epistemic uncertainties caused by mismatches between the functional form of the model and the application domain of interest
% % % model misspecification and data domain shifts
% is a primary concern of force field developers.
% The development of an accurate, computationally cheap, and easy-to-implement UQ method would thus be a valuable step towards designing automated active learning workflows and more efficient models.

While multiple UQ techniques have been applied to atomistic force fields, including dropout neural networks \cite{Wen2020}, Bayesian frameworks \cite{Angelikopoulos2012,Xie2021}, or Gaussian mixture models \cite{Zhu2023}, by far the most popular approach in practice is the use of ensemble-based methods, where the ensemble variance is taken as a measure of the predictive uncertainty of the model.
Particularly beneficial is the fact that these ensemble methods do not require any modifications to the model architecture or training algorithm, and therefore can be readily applied to any existing development workflow.
Due in large part to this simplicity and generalizability, ensemble-based UQ techniques have been widely applied towards estimating parametric uncertainty in numerous machine learning and deep learning tasks \cite{Lakshminarayanan2017,Zhou2022}, and are often considered to be the gold standard when evaluating the performance of new UQ methods for atomistic force fields \cite{Busk2021,Hirschfeld2020,Wan2021,Carrete2023,Zhu2023,Wollschlager2023}.

% Ensemble-based techniques can account for different types of uncertainty depending upon the method with which they generate the ensemble. For example, things like boosting can 

Despite their ubiquity, ensemble methods typically suffer from two main drawbacks: high computational costs (for both training and inference) and difficulties maintaining ensemble diversity that result in overconfident uncertainty estimates.
%Considering that ensemble sizes of 
% % $M = [5, 30]$ 
%models have been seen to be necessary for various practical applications of force fields \cite{Gastegger2017,Lu2023}, ensemble-based UQ methods may be infeasible for large training sets or expensive models.
In particular, ensemble sizes of approximately 5 to 30 models have been found to be necessary for various practical applications of UQ for force fields \cite{Gastegger2017,Lu2023}, making ensemble-based methods potentially infeasible for large training sets or expensive models.
Furthermore, although ensemble-based UQ techniques have been shown to typically out-perform single-model methods \cite{Tan2023}, there is growing evidence that they result in overconfident UQ measures \cite{Kahle2022,Lu2023,Fort2019,Egele2022,Zhou2022}, which is attributed to a lack of ensemble diversity.
% Similar issues have been documented in other deep learning applications, where the overconfidence is attributed to a lack of ensemble diversity \cite{Fort2019,Egele2022,Zhou2022}, which in turn stems from the methods used to generate the ensemble.
Intuitively, the most straightforward approaches to ensemble generation (training multiple models with different initial weights, data splits, or training hyperparameters \cite{Peterson2017}) will only sample models at different local minima of the loss function.
These methods therefore fail to account for the full topography of the loss surface, which has been shown in many cases to be important for minimizing generalization error \cite{hochreiter1994simplifying,Chaudhari2016,Li2017,Fort2018,Gilmer2021,Fort2020} and to be valuable for estimating model uncertainty and detecting out-of-domain data \cite{Fort2019,Fort2021}.

In this work, we develop the new \code{LTAU} method, which is an ensemble-based technique that addresses the issues of computational cost and ensemble diversity, and can be broadly applied to any model architecture.
% Critically, we outline how our method is able to circumvent the usual drawbacks of ensembles, resulting in improved ensemble diversity and only a negligible impact on inference speeds.
Specifically, we provide the following contributions:

% need to mention here how the method addresses the ''issues'' of ensembles

\begin{itemize}
    \item We outline our proposed method, \code{LTAU}, which
    utilizes the PDF of errors sampled during training
    to provide confidence intervals and expected error estimates for any prediction made with the model.
    % to provide estimates of the expected error cumulative distribution function (CDF) for any prediction made with the model.
    This use of the training error trajectories imposes no additional computational overhead during training and improves the diversity of the ensemble by incorporating additional information from various points on the loss landscape.
    \item Coupling the training PDFs to a nearest-neighbor search in the model's latent space, we show how \code{LTAU} can provide a cheap, easy-to-implement UQ measure that does not require training or evaluating multiple models. We apply \code{LTAU} to atomistic force fields (\code{LTAU-FF}), where we observe that it is 2--3 orders of magnitude faster than typical ensemble techniques.
    % , thus eliminating a major barrier to the application of ensemble-based UQ techniques with large models.
    \item We quantitatively assess the robustness of \code{LTAU-FF}, showing that it
    produces well-calibrated confidence intervals and predicts errors that correlate strongly with the true errors for data near the training domain.
    Furthermore, we observe that by using thresholds on the distance metric in the latent space, it is possible to identify out-of-domain data for which the predictions made by \code{LTAU-FF} may be less accurate.
    % outperforms the existing gold standard ensemble method on almost all tests and is able to predict a near-perfect reproduction of the distribution of test errors on an example dataset.
    \item Finally, to demonstrate the utility of \code{LTAU-FF} in practical applications, we conduct two experiments: 1) re-weighting the training set of the model to prioritize high or low uncertainty samples, and 2) applying our method to the \code{IS2RS} task from the \code{OC20} challenge (where it is infeasible to train multiple models) to predict errors in relaxation trajectories.
    % From these experiments, we observe that the UQ metric can be used to tune the training--validation performance gap and can also serve as a reliable indicator of model performance in real-world applications and on large-scale datasets.
\end{itemize}

% \begin{itemize}
%     \item We extend the notion of sample confidence \cite{Swayamdipta2020} to generic regression tasks, utilizing the cumulative distribution function (CDF) of errors observed during training to provide estimates of the likelihood that a model's predictions will fall below a chosen tolerance threshold.
%     \item By coupling the CDF to a distance-based similarity search in the model's latent space, we show that it can be used to provide efficient, accurate estimates of in-domain (ID) errors and can be readily scaled to improve calibration on out-of-domain (OOD) predictions.
%     \item As a demonstration of practical applications of our UQ metric, we conduct two experiments: 1) re-weighting the training set of the model to prioritize high/low confidence samples, and 2) applying our method to the \code{IS2RS} task from the \code{OC20} challenge to predict errors in relaxation trajectories. From these experiments, we observe that the UQ metric can be used to tune the training--validation performance gap, and can also serve as a reliable indicator of model performance in real-world applications.
% \end{itemize}

\section{Background}

The defining trait of the \code{LTAU} method developed in this work, which distinguishes it from other ensemble-based UQ methods, is its ability to
leverage information from large ensembles of models while only having to evaluate a single model during inference.
This is achieved through efficient use of logged error trajectories sampled during training and a similarity search in the model's latent space during inference.
The method builds heavily upon two key concepts in deep learning: training trajectory analysis and distance-based UQ, which are discussed further in this section.

\subsection{Training trajectory analysis}  \label{sec:traj_analysis}

Training trajectories have been used in a number of deep learning applications for the purpose of outlier detection \cite{Swayamdipta2020,Pleiss2020,Agarwal2020,Paul2021,Seedat2023}, data pruning \cite{Toneva2018,Mirzasoleiman2019,Killamsetty2020,Killamsetty2021,Sorscher2022,Seedat2023}, and influence estimation \cite{Feldman2019,Feldman2020,Guu2023}.
These approaches have leveraged the rich information that is sampled over the course of training, including loss gradients \cite{Pruthi2020,Killamsetty2020,Killamsetty2021,Mirzasoleiman2019,Paul2021,Mirzasoleiman2019,Agarwal2020}, Softmax outputs \cite{Swayamdipta2020,Rabanser2022}, or other custom metrics \cite{Pleiss2020,Toneva2018,Feldman2020,Seedat2023}.
However, many of these metrics have been designed primarily for classification tasks (i.e., they are derived using assumptions about the loss function or model architecture that are unique to classification) and require modification in order to be applied in regression settings.
In this work we show that the distribution of errors sampled during training can serve as a useful extension of these trajectory analysis techniques to regression tasks for estimating parametric uncertainty and detecting out-of-domain samples.

\subsection{Distance-based UQ}
\label{sec:distance_uq}

A critical assumption of the method developed in this work is that points which are close to each other in the model's latent space also have similar error PDFs.
% , which is a concept that is discussed further in \App{supp:scaling}.
On a smoothly-varying manifold, proximity between points is often used to infer or model function values.
This notion has been successfully applied to develop tools like radial basis functions, Gaussian process regression, and many others.
Similarly, this concept has been applied to UQ, where a simple method for estimating the uncertainty in a model's prediction for a test point is to use the weighted Euclidean distance (in either the input space or a latent space) between the test point and the nearest training point \cite{Liu2018,Abdar2021,Sorscher2022}.

% The rationale behind distance-based UQ metrics is that a model is less likely to have accurate predictions on points which are further from the training set.
While related methods have been used for UQ with atomistic force fields \cite{Janet2019,Pernot2022,Zhu2023}, a distance metric alone is insufficient for constructing well-calibrated uncertainty estimates, and usually requires additional calibration techniques or making assumptions regarding the distribution of errors \cite{Hu2022}.
Specifically, any distance-based UQ metric would be expected to fail in the case where a point is far enough from any other point that the assumption of local smoothness of the model uncertainty does not hold.
This issue is discussed further in \Sec{calibration}
% and \App{supp:scaling}
, where we find that
the confidence intervals and expected errors predicted by \code{LTAU-FF} become less accurate when used on out-of-domain (OOD) data.
% the uncertainties provided by \code{LTAU-FF} must be scaled in order to improve their calibration to out-of-domain data.
This issue is not unique to our method, and emphasizes the importance of coupling UQ metrics with suitable tools for detecting OOD data.
% in fact we show that a scaled version of \code{LTAU-FF} still out-performs a scaled version of a typical ensemble-based UQ method.
% In our analysis, we observe that \code{LTAU-FF} can be calibrated using a simple scaling factor, which will be discussed further in \Sec{calibration}.
% We use a particular calibration technique described in \Sec{calibration}.
% In this work, we couple the CDFs of errors on the training points to a distance-based similarity search for predicting uncertainties on test points.
% We use the \code{FAISS} software package \cite{johnson2019billion} to perform efficient similarity searches, enabling low-cost UQ relative to the computational cost of a forward pass of the model.

\section{Methods}\label{sec:methods}

\subsection{Model} \label{sec:model}

We use the NequIP model \cite{Batzner2022} as the deep learning force field in this work.
NequIP is a graph-based message-passing neural network that uses spherical harmonics to represent local atomic environments with equivariant features.
The use of message-passing and an equivariant architecture, respectively, have been shown to be useful for incorporating long-range interactions in the model and avoiding issues caused by the inability of invariant models to distinguish between symmetrically-equivalent atomic environments \cite{Batatia2022b}.
Because of this, NequIP (alongside other equivariant message-passing models) has achieved state-of-the-art performance for a number of applications \cite{Schutt2021,Reiser2022}.
The hyperparameters of the model are adjusted for each dataset, with details provided in \App{supp:hyperparams}.

\subsection{Datasets} \label{sec:datasets}

Three training datasets are used in this work: \code{3BPA} \cite{Kovcs2021}, \code{Carbon\_GAP\_20} \cite{Rowe2020}, and the \code{200k} split of the \code{S2EF} task from the \code{OC20} challenge \cite{Chanussot2021}.
These datasets were chosen deliberately to provide systematic tests of the limitations of our method with increasingly complex data.
We start with a simple molecular test case for refining our method (\code{3BPA}), then expand to a more diverse dataset typical of solid-state force field fitting tasks (\code{Carbon\_GAP\_20}), and finally test against a challenging real-world application (\code{IS2RS} task from \code{OC20}) that is intractable using typical ensemble methods.
Detailed descriptions of the datasets are found in \App{supp:data} and the referenced citations.
All datasets are publicly available from their original sources.

\subsection{Training details}
\label{sec:training_details}

For the most part, standard published hyperparameters and techniques were used for training the NequIP models in this work, with details provided in \App{supp:hyperparams}.
As a notable exception, for all analysis shown in this work for the \code{3BPA} and \code{Carbon\_GAP\_20} datasets we trained the models \textbf{only to forces}, which is in contrast to typical force field fitting workflows that usually also include an energy term in the loss function.
We chose to weight the energy contributions to 0.0 in these cases for two reasons: 1) because a system's energy cannot be uniquely decomposed into a per-atom quantity, thus making it a less suitable target for constructing a per-atom UQ metric; and 2) because a trade-off is commonly observed in practice between the energy and force terms in the loss function, raising concerns that training to energies could interfere with our analysis.
Training only to forces for the \code{3BPA} and \code{Carbon\_GAP\_20} datasets thus allowed us to obtain a clearer understanding of the performance of our method.
% We chose to weight the energy contributions to 0.0 in these cases to remove any possible spurious correlations caused by coupling between the energy and force terms in the loss function, which may have interfered with the UQ analysis.
For the \code{OC20} dataset, we weighted the energy term to 1.0 (and the force term to 100.0) to help avoid unexpected behavior during energy minimization; we note that this still resulted in strong correlation between the UQ metric provided by \code{LTAU-FF} and the accuracy of the predicted ground-state structure, as shown in \Sec{oc20}.
% we note that this did not appear to affect the performance of our method.
% For \code{Carbon\_GAP\_20} a random 90-10 training--validation split was used; for \code{3BPA} the predefined training set was used with no validation set.
% Batch sizes were taken to be 1 (\code{Carbon\_GAP\_20}) or 5 (\code{3BPA}).
% To provide some rudimentary details about the training settings: we used the AMSGrad variant of the Adam optimizer with an initial learning rate of 0.005, the \code{ReduceLROnPlateau} scheduler with a patience of 10, 

\subsection{LTAU--FF} \label{sec:ltau}

As the primary contribution of this paper, we propose the \textbf{L}oss \textbf{T}rajectory \textbf{A}nalysis for \textbf{U}ncertainty (\code{LTAU}) method (\Alg{ltau_p} and \Alg{ltau_test}), and demonstrate its application to atomistic force fields (\code{LTAU-FF}: \code{LTAU} for atomistic \textbf{F}orce \textbf{F}ields).
The \code{LTAU} method builds upon the concepts described in \Sec{intro}, using the distributions of errors observed by the ensemble of models sampled during training to approximate the PDFs of all training points.
The training error PDFs are then coupled to a distance-based similarity search for estimating the PDFs for test predictions.
% using the errors of the ensemble of models sampled during training to approximate the CDF, then coupling the approximated CDF to a distance-based similarity search for estimating uncertainty on test predictions.
% By leveraging the error information obtained during training, and the efficiency of the similarity search methods in \code{FAISS}, \code{LTAU} provides a cheap and easy-to-implement avenue for obtaining uncertainty estimates for any regression task.
Implementing \code{LTAU} in an existing workflow will typically only require patching the training code to log per-sample errors at every epoch and adding some basic post-processing steps. We provide code and utility functions for performing this postprocessing and working with the models presented in this work,
which will be released upon completion of the peer review of this article.
% at \url{https://lc.llnl.gov/gitlab/vita1/LTAU--FF}.

% Central to the \code{LTAU} technique is the notion of sample confidence, as described in \Sec{traj_analysis}.
% In \cite{Swayamdipta2020}, sample confidence is defined as the mean probability of the correct class label being predicted across training epochs and is computed using the outputs of the final Softmax layer.
% Since regression tasks do not normally use a Softmax output layer,
% , as is available for classification tasks, and there is no binary notion of ``correctness''
% To make \code{LTAU} as easy to implement and computationally efficient as possible, we choose to use the per-sample PDFs of errors observed during training to describe the uncertainty of the model.
% and approximate the uncertainty on test points.
% estimate the model's errors and produce confidence intervals for its predictions.
% The estimated PDF can then be used to define the ``confidence'' in a given prediction as the likelihood of the prediction having errors below a chosen tolerance threshold.
% define the model's confidence at any given tolerance threshold.
% evaluated at a chosen tolerance threshold, $atol$, as the measure of confidence.
% In this sense, a sample with high confidence at a given tolerance threshold is one which has a high probability of having errors below the threshold over the course of training.
For \code{LTAU-FF}, the first step to computing the
% sample confidence
uncertainty
in this manner is to train a model and log the errors at every epoch (\Alg{ltau_p}). By logging the model's errors on every sample (i.e., every atom) at every epoch, we can construct error trajectories $T_i = \{ \epsilon_i^1, \cdots, \epsilon_i^E \}$ for every atom $i$, where $\epsilon_i^t = |\textbf{F}_{i,DFT}-\textbf{F}_i^t|_2$ is the $L_2$ norm of the force error vector between the true ($\textbf{F}_{i,DFT}$) and predicted ($\textbf{F}_i^t$) forces for atom $i$ at the end of epoch $t$, and $E$ is the total number of training epochs.
% We note that instead of using the $L_2$ norm for computing $\epsilon_i$, each Cartesian coordinate could be treated individually, yielding $\epsilon_{ic}$ (with $c$ indexing the Cartesian direction).
% This would allow uncertainties to be obtained along each axis, but would incur higher memory costs when logging the per-sample errors.
Alternative error metrics can also be used instead of the $L_2$ norm -- for example, the mass-normalized errors explored in \cite{Wang2023} would be particularly useful for multi-element systems.

\SetKwComment{Comment}{// }{}

\begin{minipage}[t]{0.5\linewidth}
\begin{algorithm}[H]
    \SetKwInOut{Input}{Input}
    \SetKwInOut{Output}{Output}
    
    \Input{Model, $\mathcal{F}$; training set $\mathcal{S}$ of size $N$; epochs, $E$; bin edges, $bins$}
    \Output{PDFs of training points, $p = \{ \text{PDF}_1, \ldots, \text{PDF}_N \}$; descriptors of training points, $D$}
    \BlankLine
    \For{$e = 1 \ldots E$} {
        Train($\mathcal{F}$)
        
        Log(Error$(\mathcal{F}, \mathcal{S})$, $e$)
    }
    
    % \tcp*[h]{load ExN matrix of trajectories}
    $T \gets $ LoadErrorTrajectories()
    
    % \tcp*[h]{compute column-wise histograms}
    
    $p \gets $ Histograms($T, bins$)% / $N$
    
    $D \gets $ Descriptors($\mathcal{F}$, $\mathcal{S}$)% \Comment*[r]{extract the descriptors for all training points}
    
    return $p$, $D$
    
    \caption{Compute PDFs and descriptors of training points}
    \label{alg:ltau_p}
\end{algorithm}
\end{minipage}
\begin{minipage}[t]{0.5\linewidth}
\begin{algorithm}[H]
    \SetKwInOut{Input}{Input}
    \SetKwInOut{Output}{Output}

    \Input{Model, $\mathcal{F}$; training PDFs, $p$; training descriptors, $D$; test point, $x_j$; number of nearest neighbors, $k$}
    \Output{Estimated PDF of test point, $p_j$}
    \BlankLine
    % \texttt{\\}
    
    $d_j \gets $ Descriptor($\mathcal{F}$, $x_j$)% \Comment*[r]{extract the descriptor for the test point}

    % \texttt{\\}

    $\mathcal{N}_j \gets $ NearestNeighbors($d_j$, $D$, $k$)% \Comment*[r]{find the indices of the $k$ nearest neighbors of $d_j$ in $D$}
    
    return $\frac1k \sum_{i \in \mathcal{N}_j} p_i$% \Comment*[r]{return the average uncertainty of the $k$ nearest neighbors}
    
    % \BlankLine
    
    \caption{Estimate PDF of test point}
    \label{alg:ltau_test}
\end{algorithm}
\end{minipage}

Once training is complete, as a post-processing step we can compute the error PDFs, $p_i$, for each atom $i$ by loading the array of logged error trajectories, then binning the observed errors into a chosen set of bins.
% (logarithmically-spaced bins are recommended).
Though a maximum error value, $\epsilon_{max}$, defining the upper edge of the uppermost bin may not necessarily be known \textit{a priori}, there are many reasonable choices that could be made without any knowledge of the expected test errors.
For example, $\epsilon_{max}$ could be set to be a multiple of the maximum error observed during training, or the maximum ``acceptable'' error as determined by the application of interest.
In \Sec{calibration}, we chose to set $\epsilon_{max}$ to be the maximum error observed during training.
To use the PDFs computed on the training set for predicting the PDF of a test point, $j$, we compute $p_j = \frac1k \sum_{i \in \mathcal{N}_j} p_i$ by averaging the value in each bin over $\mathcal{N}_j$, the $k$ nearest samples of $j$ from the training set (\Alg{ltau_test}).
% , the average of the PDFs for the nearest $k$ samples, $\mathcal{N}_j$, from the training set (\Alg{ltau_test}).
In practice, we observed that \code{LTAU} was relatively insensitive to the choice of $k$, so we chose to use $k=10$ for all experiments.
% In practice, a reasonable choice of $k$ can be made by taking the smallest value which maximizes the calibration of the model to the training set (see \Fig{supp:k_convergence}).
% However, we note that in our experiments \code{LTAU-FF} was relatively insensitive to the choice of $k$.

The $k$ nearest neighbors are obtained by performing a similarity search in the latent space of the model.
Specifically, using the output of the last message-passing layer of the NequIP model (prior to the linear readout layers) as the atomic descriptor, as is done elsewhere in the literature \cite{Zhu2023}.
This means that we must evaluate the model for the full training set, store the latent descriptors for all training atoms, then compare the descriptors for each test atom, $j$.
The cost of computing $p_j$ then necessarily scales with the size of the training set, the number of neighbors $k$, and the dimensionality of the latent space (which was 32 for all experiments in this work).
Note that the latent descriptors will already be computed when performing a forward pass of the model, so $p_j$ can be efficiently estimated with only a minor increase in cost compared to a force evaluation, as shown in \Sec{computational_cost}.

Since typical training sets for force fields can include $\mathcal{O}(10^4)$ - $\mathcal{O}(10^7)$ atoms, we rely upon the \code{FAISS} \cite{johnson2019billion} package for efficiently searching for the $k$ nearest neighbors of test points.
In the case of the the \code{Carbon\_GAP\_20} and \code{OC20} datasets, where the training sets were particularly large, we used the \code{IndexHNSWFlat} algorithm provided by \code{FAISS}.
\code{IndexHNSWFlat} is an \textit{approximate} nearest neighbor search that scales well to large datasets while maintaining good accuracy.
For the \code{3BPA} dataset, an exact brute-force algorithm was used.
% Depending upon the size of the training set, it may be useful to use only \textit{approximate} nearest neighbor search methods, which are discussed briefly in \Sec{computational_cost}.
% For more thorough discussion of approximate nearest neighbor searches, we refer the reader to the \code{FAISS} documentation and associated publication \cite{johnson2019billion}.
Additional details regarding the \code{FAISS} hyperparameters used in this work can be found in \App{supp:hyperparams}.

\section{Results and Discussion}

\subsection{Comparisons to an ensemble-based method}
\label{sec:calibration}

% Mention improving sharpness by using element-specific or atom-specific atol values

As an initial comparison between \code{LTAU-FF} and a typical ensemble method (henceforth referred to as \code{Ensemble}), we train an ensemble of 10 models to the \code{3BPA} and \code{Carbon\_GAP\_20} datasets, then compare the quality of the uncertainty estimates provided by each method on the test sets.
% using tools and metrics from the literature \cite{Tran2020,Zhu2023}.

% then compare the ability of each method to estimate the ensemble errors on force predictions.
In \Fig{parity_cal_3bpa}a and \Fig{parity_cal_gap20}a we plot the per-atom force uncertainties versus the true force errors observed on the test sets (averaged over the ensemble).
The uncertainty predicted by \code{LTAU-FF} for atom $i$ is obtained by taking the expectation value of $p_i$ (the PDF of errors for atom $i$).
As is typically done in the literature \cite{Busk2021,Hirschfeld2020,Wan2021,Carrete2023,Zhu2023,Wollschlager2023}, for the uncertainty predicted by \code{Ensemble} we use the standard deviation of the ensemble's force predictions averaged over the Cartesian directions.
Note that the uncertainty metric provided by \code{LTAU-FF} can be directly interpreted as an error estimate, whereas those obtained from \code{Ensemble} cannot since they simply correspond to the spread in predictions by different models in the ensemble.

% The calibration curves \cite{Tran2020} in \Fig{parity_cal_3bpa}b and \Fig{parity_cal_gap20}b show how well the uncertainty
% As an initial test of the ability of \code{LTAU-FF} to estimate errors on force predictions, we trainensembles of models to the \code{3BPA} and \code{Carbon\_GAP\_20} datasets, then quantitatively compare the calibration curves of the model on subsets of the test data, as is done in \cite{Tran2020}.

To generate the calibration curves shown in \Fig{parity_cal_3bpa}b and \Fig{parity_cal_gap20}b, we compute the fraction of points whose true errors fall within confidence intervals ranging from 0\% to 100\% for each method.
The calibration curves can thus be thought of as a comparison between the true and predicted CDF of the test errors, where a well-calibrated model would have a calibration curve that falls close to the $x=y$ line \cite{Tran2020}.
For \code{Ensemble}, the error thresholds corresponding to a given confidence interval can be obtained by assuming that the errors of each atom follow a Gaussian distribution with variance equal to the ensemble variance of the force predictions for that atom, then computing the z-score at the desired confidence level.
For \code{LTAU-FF}, since the CDFs can be directly computed from the predicted PDFs, the error thresholds can be determined by finding the bin edge at which the CDF first reaches the desired confidence level.

\Tab{metrics} provides a quantitative analysis of \Fig{parity_cal_3bpa} and \Fig{parity_cal_gap20}.
The Pearson ($\mathcal{C}_P$) and Spearman ($\mathcal{C}_S$) coefficients are used to measure the correlation between the uncertainty estimates and the true errors.
We quantify the degree of miscalibration by computing the area ($|\mathcal{A}|$) between the calibration curves and the ideal $x=y$ line, as is done in \cite{Tran2020}.
To provide additional analysis, we further decompose the miscalibration area into the area below/above the $x=y$ line ($\mathcal{A}^+$ and $\mathcal{A}^-$, denoting the degree of \textit{over}confidence and \textit{under}confidence of the method, respectively).
Finally, the ``sharpness'' \cite{Tran2020} is calculated by computing the variance of each atom's predicted error distribution, averaging over all atoms in the test set, then taking the square root (to recover the original units of eV/\AA).
% as the square root of the average over the test sets of the variance of the uncertainty estimates.
An ideal UQ method would have uncertainty estimates which strongly correlate with the true errors (high $\mathcal{C}_P$ and/or $\mathcal{C}_S$) and have confidence intervals which are as small as possible (low $s$) while still capturing the true errors (low $\mathcal{A}^+$, $\mathcal{A}^-$, and $|\mathcal{A}|$).
For the \code{LTAU-FF} results in \Tab{metrics}, the metrics are computed for each model in the ensemble individually, then averaged over the ensemble. 
The variance in the metrics for \code{LTAU-FF} was negligible (less than 0.01 in all cases).
For \code{Ensemble}, the uncertainty measure is the same for all models in the ensemble, so no averaging is possible.

\subsubsection*{\code{3BPA}}  \label{sec:calibration_3bpa}

\begin{figure}[ht!]
\centering
\includegraphics[width=0.95\linewidth]{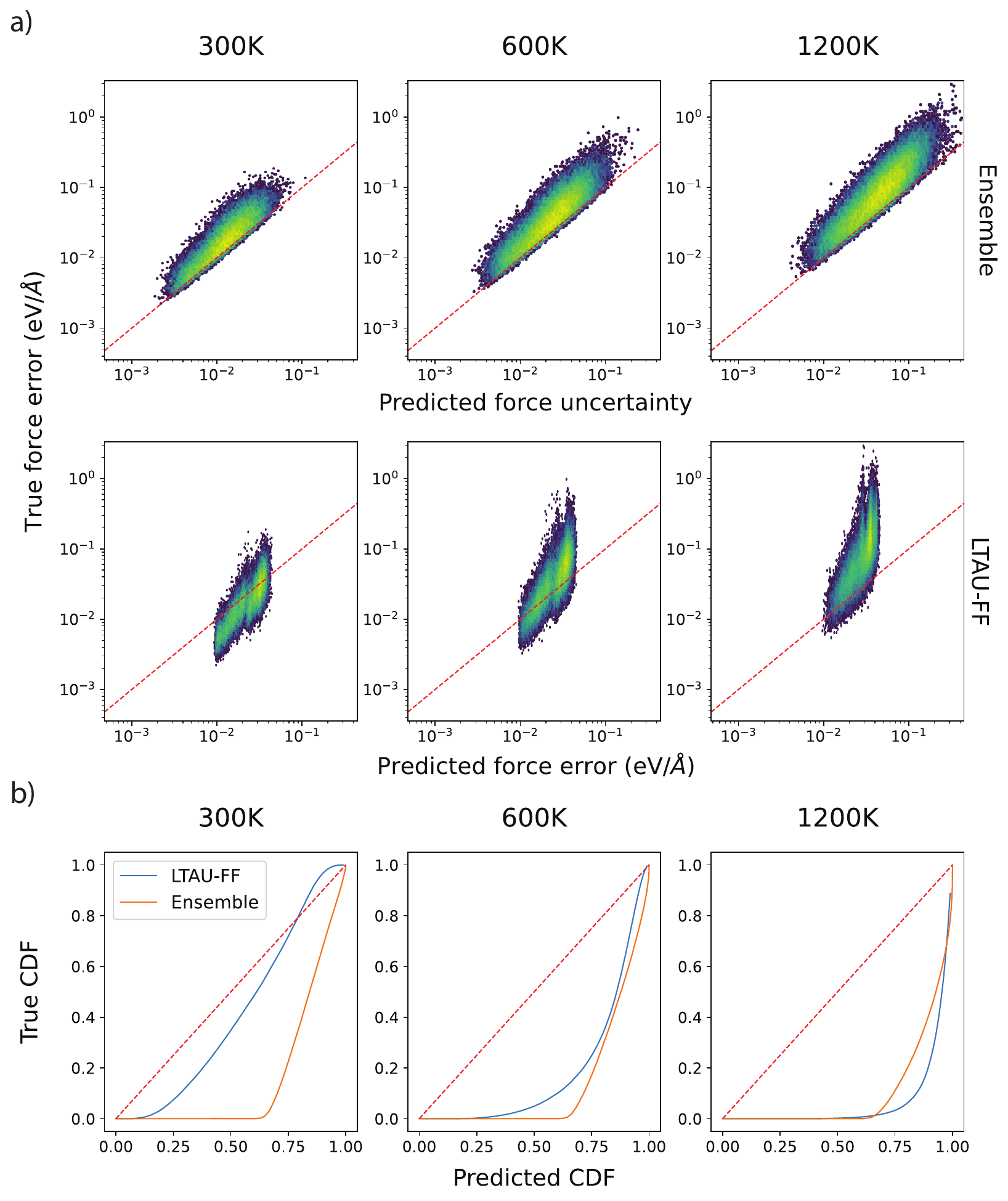}
\caption{
Comparison of \code{LTAU-FF} and \code{Ensemble} on the the \code{3BPA} test sets.
Panel \textbf{a} compares the uncertainty estimates for each method to the true errors observed on the test sets (note the log-scaled axes), and can be used to obtain $\mathcal{C}_P$ and $\mathcal{C}_S$ in \Tab{metrics}.
The calibration curves, as described in \cite{Tran2020}, show how well the predicted confidence intervals capture the true errors (with an ideal model falling on the $x=y$ line), and can be used to obtain $\mathcal{A}^+$, $\mathcal{A}^-$, $|\mathcal{A}|$, and $s$ in \Tab{metrics}.
% Panel \textbf{a} shows the parity plots for each method and test set (note the log-scaled axes), while panel \textbf{b} shows the calibration curves as described in \Sec{calibration}.
}
\label{fig:parity_cal_3bpa}
\end{figure}

Before analyzing the results in \Fig{parity_cal_3bpa} and \Tab{metrics}, it is important to recall some of the main features and limitations of \code{LTAU-FF}.
First, because \code{LTAU-FF} is also an ensemble method, we can expect that it should perform similarly to \code{Ensemble}, though with the additional benefit of having more conservative uncertainty estimates due to the increased diversity of the ensemble sampled during training.
Second, since \code{LTAU-FF} relies upon a distance-based similarity search for estimating the PDFs of errors for test points,
it can be expected to suffer from the drawbacks described in \Sec{distance_uq}, leading to decreased performance when the test data are far from the training set.
% it can be expected that its error estimates may be less accurate on data which are far from the training set.

% The effects of both of these features are readily apparent in \Fig{parity_cal_3bpa} and \Tab{metrics}, where \code{LTAU-FF} outperforms \code{Ensemble} on the in-domain (ID) data, but shows decreasing accuracy on the out-of-domain (OOD) data.

% The results shown in \Fig{parity_cal_3bpa} and \Tab{metrics} support these hypotheses, where \code{LTAU-FF} is better calibrated than \code{Ensemble} for the in-domain (ID) data, but 

Considering first only the 300K data
% in \Fig{parity_cal_3bpa} and \Tab{metrics}
, it can be seen that \code{LTAU-FF} outperforms \code{Ensemble} for the in-domain (ID) data.
Both methods have reasonably good $\mathcal{C}_P$ and $\mathcal{C}_S$ coefficients, but \code{LTAU-FF} has an $|\mathcal{A}|$ area that is $3\times$ lower and a sharpness that is $2\times$ lower than those of \code{Ensemble}. 
This is consistent with our expectations that the increased diversity of the training ensemble would help address the documented overconfidence issues of \code{Ensemble} \cite{Kahle2022,Lu2023}.
% While the correlation coefficients are slightly higher for \code{Ensemble}, we argue that the drastically improved calibration and lower sharpness of \code{LTAU-FF}

However, when moving to the higher-temperature (600K and 1200K) data, we observe decreased performance from \code{LTAU-FF} quantified by lower correlation coefficients and higher miscalibration areas.
We hypothesize that this is caused by the fact that the nearest-neighbor search used by \code{LTAU-FF} is unable to account for the effects of increasing distance in the latent space when predicting the PDF of errors for test points.
For example, even if one the $k$ nearest-neighbors is significantly further away from a test point $j$ than another of the $k$ nearest-neighbors, it will still have an equal influence on the uncertainty estimate.
This deficiency is one of the most important weaknesses of \code{LTAU-FF}.

We propose a possible solution to this issue in the next section using distance thresholds to detect out-of-domain (OOD) data; however, we emphasize that further improvements are still necessary.
% This deficiency is one of the most important weaknesses of \code{LTAU-FF} that we believe must be addressed in future work, though we discuss a possible solution in the next section.
Modifying \code{LTAU-FF} to improve its performance on OOD or ``near OOD'' data (that which is OOD, but with distributions similar to the training data) \cite{Winkens2020}, for example by weighting the neighbor average based on neighbor distance or coupling it to additional OOD-detection algorithms, would greatly improve the method.
Furthermore, it is worth noting that both \code{Ensemble} and \code{LTAU-FF}, like most UQ metrics, would benefit from additional re-calibration techniques such as the use of an external ``calibration set'' \cite{Kuleshov2018}.
Especially considering the high $\mathcal{C}_P$ coefficients, even a simple linear transformation of the uncertainty estimates would likely greatly improve the calibrations of both methods.
Nevertheless, the good calibration of \code{LTAU-FF} on ID data \textit{without} requiring additional calibration is a notable strength of the method.

\subsubsection*{\code{Carbon\_GAP\_20}}

\begin{figure}[ht!]
\centering
\includegraphics[width=\linewidth]{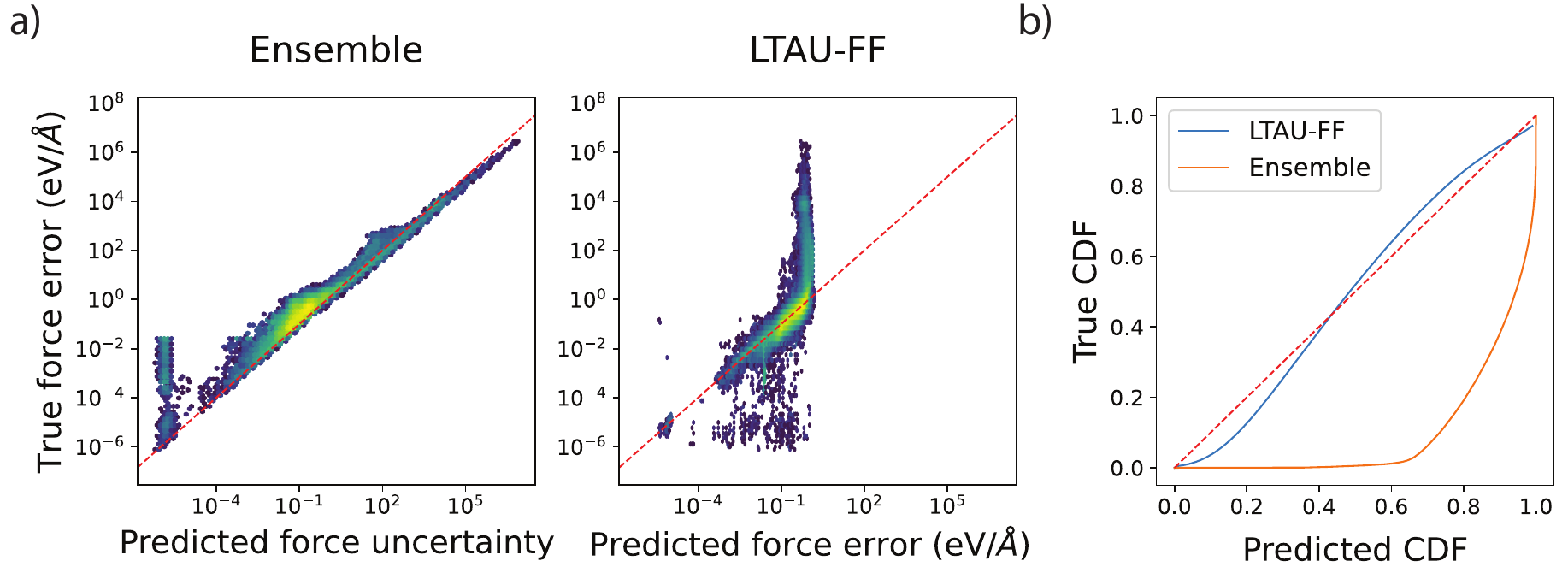}
\caption{
Comparison of \code{LTAU-FF} and \code{Ensemble} on the \code{Carbon\_GAP\_20} test set.
Predicted uncertainty vs. true force errors shown in panel \textbf{a} and calibration curves in panel \textbf{b} are as described in \Fig{parity_cal_3bpa}.
Quantitative analysis is provided in \Tab{metrics}.\
The ``wall'' of points around 1 eV/\AA~ where \code{LTAU-FF} makes poor predictions are those which we identify as being OOD based on their nearest-neighbor distance in the latent space (see \App{supp:ood}).
% For the parity plots in panel \textbf{a}, values were clipped to the range of errors observed during model training. This was done to remove outliers with unreasonably high errors (on the order of $1e8$) that we believe were caused by poorly converged simulations when generating the ground-truth values.
}
\label{fig:parity_cal_gap20}
\end{figure}

The results on the \code{Carbon\_GAP\_20} dataset shown in \Fig{parity_cal_gap20} reinforce the conclusions drawn from the \code{3BPA} results: that \code{LTAU-FF}  outperforms \code{Ensemble} on ID data and is generally the better calibrated metric, but may struggle
to estimate errors for OOD data.
Most notably, the results for \code{LTAU-FF} in \Fig{parity_cal_gap20}a appear to hit a ``wall'' where the method does not predict errors larger than approximately 1 eV/\AA.
A similar effect can be seen for the \code{3BPA} results in \Fig{parity_cal_3bpa}a around 0.05 eV/\AA.
% The cause of this behavior can be understood by recalling the choice described in \Sec{ltau} of setting $\epsilon_{max}$ (the upper bound of the bins used for constructing the training PDFs) to be the maximum error observed during training, which in the case of \code{Carbon\_GAP\_20} .
% Again, \code{LTAU-FF} is well-calibrated even without any additional re-calibration techniques, and shows a strong $\mathcal{C}_S$ correlation with the true errors.
% However, in \Fig{parity_cal_gap20}, 
% to accurately estimate errors which are significantly larger than those seen during training.

While this effect is in part due to the fact that the errors observed during training may be significantly smaller than those observed on the test sets, 
% This effect is a direct result of the issues described in \Sec{distance_uq} related to distance-based UQ metrics, and
the issue
can be best understood by analyzing the distributions of nearest-neighbor distances in \Fig{supp:distances}.
It can be seen that for both datasets there are a
small number of points in the training data which have a significantly larger first nearest-neighbor distance than the rest of the training set.
% sharp increase in distances for a small portion of the training data, which we hypothesize correspond to outliers and OOD data.
By choosing a distance cutoff close to the
lower bound of these higher distances,
% onset of this increase,
we can define a point as being ID if it has a nearest neighbor distance below the cutoff, or OOD otherwise.
We observe in \Fig{supp:3bpa_ood} and \Fig{supp:gap20_ood} that the ``walls'' in \Fig{parity_cal_3bpa} and \Fig{parity_cal_gap20} roughly correspond to the data which are identified as being OOD based on the distance cutoffs.
This observation suggests a possible avenue forward for improving \code{LTAU-FF} to warn when the method may be unreliable on OOD data, thus hopefully addressing the main limitation of this work.

\begin{table}[ht]
\centering
\caption{Metrics for comparing UQ methods. $\mathcal{C}_P$ and $\mathcal{C}_S$ are the Pearson and Spearman correlation coefficients; $\mathcal{A}^+$, $\mathcal{A}^-$, and $|\mathcal{A}|$
% describe the area between the calibration curves and the line $x=y$,
provide measures of the degree of miscalibration \cite{Tran2020} (overconfidence, underconfidence, and total, respectively); $s$ is the ``sharpness,'' \cite{Tran2020} which measures how narrow the confidence intervals are (in the same units as the model predictions). Arrows ($\uparrow$, $\downarrow$) denote if larger or smaller values are better, respectively.
The near-zero $\mathcal{C}_P$ score of \code{LTAU-FF} on the \code{Carbon\_GAP\_20} dataset is caused by the ``wall'' in \Fig{parity_cal_gap20}.
% Bolded values are used to emphasize the best values for each test set.
% The $s$ value for \code{Ensemble} on \code{Carbon\_GAP\_20} were computed after removing outliers, as described in \Fig{parity_cal_gap20}.
}
\label{tab:metrics}
\begin{tabular}{llcccccc}
    \hline \hline
    Test set & Method & $\mathcal{C}_P (\uparrow)$ & $\mathcal{C}_S (\uparrow)$ & $\mathcal{A}^+ (\downarrow)$ & $\mathcal{A}^- (\downarrow)$ & $|\mathcal{A}| (\downarrow)$ & $s (\downarrow \text{, eV/\AA})$ \\
    \hline
     
    \multirow{2}{7em}{\code{3BPA} (300K)} & \code{Ensemble} & \textbf{0.83} & \textbf{0.90} & 0.34 & \textbf{0.0} & 0.34 & 0.12 \\
    & \code{LTAU-FF} & 0.73 & 0.84 & \textbf{0.10} & 0.01 & \textbf{0.11} & \textbf{0.06} \\
    \hline
    \multirow{2}{7em}{\code{3BPA} (600K)} & \code{Ensemble} & \textbf{0.81} & \textbf{0.89} & 0.36 & 0.0 & 0.36 & 0.16 \\
    & \code{LTAU-FF} & 0.63 & 0.82 & \textbf{0.31} & 0.0 & \textbf{0.31} & \textbf{0.07} \\
    \hline
    \multirow{2}{7em}{\code{3BPA} (1200K)} & \code{Ensemble} & \textbf{0.80} & \textbf{0.89} & \textbf{0.40} & 0.0 & \textbf{0.40} & 0.23 \\
    & \code{LTAU-FF} & 0.54 & 0.77 & 0.43 & 0.0 & 0.43 & \textbf{0.08} \\
    \hline
    \multirow{2}{7em}{\code{Carbon\_GAP\_20}} & \code{Ensemble} & \textbf{0.99} & 0.87 & 0.40 & \textbf{0.0} & 0.40 & 12.11 \\
    & \code{LTAU-FF} & 0.02 & \textbf{0.91} & \textbf{0.02} & 0.02 & \textbf{0.04} & \textbf{0.29} \\
    
    \hline \hline
\end{tabular}
\end{table}

% \begin{figure}
% \centering
% \includegraphics[scale=0.4]{figures/main/gap20_pdf_total.pdf}
% \label{fig:gap20_pdf_total}
% \caption{Probability distribution functions for the \code{Carbon\_GAP\_20} test set. The true distribution was obtained by averaging the errors from an ensemble of models. The \code{Ensemble} and \code{LTAU-FF} are predicted distributions using the corresponding methods, as described in \Fig{calibration_3bpa} and \Sec{calibration}. We hypothesize that the peak in the \code{LTAU-FF} predictions around 0.02 corresponds to a large number of graphite-like structures where the model was unable to learn the long-range interactions between layers of graphene (see \Fig{supp:weird_bin}, which confirms that the bin between 0.02 and 0.03 contains many graphite-like structures).}
% \end{figure}

\subsection{Applications}
\label{sec:applications}

\subsubsection*{Tuning the training--validation gap}

As an initial application of our method, we test if the uncertainty estimates provided by \code{LTAU-FF} can be used to tune the training-validation gap on the \code{Carbon\_GAP\_20} dataset.
\Fig{supp:training_curves} shows that by increasing the weight of high-uncertainty samples it is possible to decrease the training error (while correspondingly increasing the validation error).
Conversely, by increasing the weight of the low-uncertainty samples, it is possible to \textit{decrease} the training-validation gap, which would be expected to result in a more generalizable model.
We further confirmed the performance gaps by computing the ensemble test errors for each weighting scheme, as shown in \Fig{supp:upweight_errors} where increasing the weight on low-uncertainty points decreased the errors on outlying test points by an order of magnitude.
Additional details for this experiment are provided in \App{supp:upweight}.

These results are consistent with literature from other fields of deep learning where it has been shown that measures of sample difficulty may be valuable for tuning model performance.
For example, Feldman et al. \cite{Feldman2019,Feldman2020} hypothesized that difficult (high uncertainty) samples are essential for fitting to datasets with long-tailed distributions, which is supported by our observation that increasing the weight of high-uncertainty samples leads to increasing degrees of overfitting.
This connection to the literature implies many possible applications of \code{LTAU-FF} that build upon the work by Feldman et al. for dataset pruning and active learning, where it has been shown that sample difficulty metrics and training trajectory analysis can be used for constructing optimal training sets \cite{Toneva2018,Swayamdipta2020,Paul2021,Sorscher2022}

\subsubsection*{Leveraging the UQ metric during simulations}
\label{sec:oc20}

The final test of \code{LTAU-FF} that we performed was to train a model to the \code{200k} split of the \code{OC20 S2EF} task, then to see if \code{LTAU-FF} could predict model performance on the \code{IS2RS} task.
Due to the size of the \code{OC20} dataset (over 14 million data points in the \code{200k} training split), it is not computationally feasible to train an ensemble of models, making \code{LTAU-FF} a particularly attractive alternative to \code{Ensemble} for this task.
In the \code{IS2RS} task, a model is given an initial atomic configuration and is asked to predict the relaxed state, which is a broadly-applicable and extremely valuable task for materials discovery.
While some models use the ``direct'' approach of predicting the relaxed structure directly from the input configuration, an alternative approach (shown to achieve better performance in \cite{Chanussot2021_corrected}) is to perform relaxation via energy minimization, which is our approach in this work.

\begin{figure*}[ht!]
\centering
\includegraphics[width=0.6\linewidth]{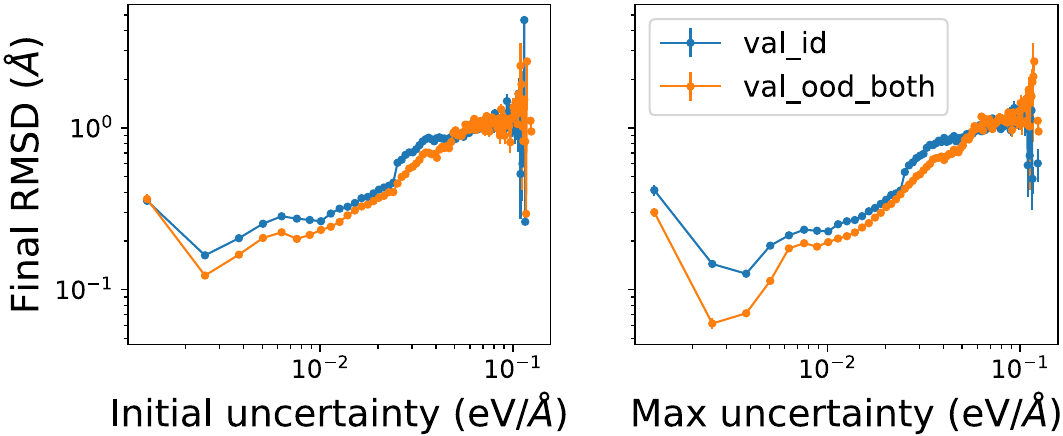}
\caption{RMSD between atoms of DFT-relaxed and model-relaxed samples from the \code{IS2RS} task of \code{OC20}, binned by predicted uncertainty. Panels correspond to different choices of snapshots along the relaxation trajectory to use for predicting the final RMSD.
The splits identified by \cite{Chanussot2021} as being in-domain (\code{val\_id}) or out-of-domain (\code{val\_ood\_both}) are shown in blue or orange, respectively.
% For all panels except the third (``Max UQ''), we observe an initial plateau for uncertainties below approximately $0.3$, then a linear increase in RMSD with increasing uncertainty.
% Only the ``surface'' and ``adsorbate'' atoms are considered in this figure, since the ``bulk'' atoms are held fixed during energy minimization following the practices outlined in the \code{OC20 IS2RS} task.
Distances are averaged within each bin, and error bars correspond to the standard error for each bin.
% Note that \cite{Chanussot2021} uses a maximum RMSD value of 0.5 \AA~ when computing the ``Average Distance within Threshold'' (ADwT) metric on the \code{IS2RS} task.
Only the adsorbate and surface atoms were considered in these plots, as is done in \cite{Chanussot2021}.
}
\label{fig:oc20}
\end{figure*}

A particularly challenging aspect of predicting errors from a relaxation trajectory is that relatively small errors at any point in the trajectory may drastically alter the predicted relaxed structure by causing the model to relax towards an incorrect local energy minimum.
Because of this, in \Fig{oc20} we consider both the initial uncertainty as well as the maximum uncertainty predicted during energy minimization, both of which show strong correlation with the final root mean-squared displacement (RMSD) of the relaxed structure.
Notably, there is significantly more variation in the RMSD at larger uncertainties, reflecting the fact that
sufficiently large errors at any point in the simulation can drastically alter the relaxation trajectory.
% it is inherently difficult to predict simulation failure using a single snapshot along the trajectory.
These results motivate the importance of further research into uncertainty propagation methods for molecular simulations, which would benefit greatly from the computational efficiency of \code{LTAU-FF} to allow the methods to be applied to large-scale simulations.

\subsection{Computational cost}  \label{sec:computational_cost}

To better understand the computational cost of \code{LTAU-FF} relative to other aspects of model evaluation, we ran profiling tests computing the atomic forces and uncertainties for 500 randomly sampled atomic configurations from each test set.
These results were obtained using the CPU implementation of FAISS and a \code{TorchScript}-compiled NequIP model running on a single NVIDIA V100 GPU.
The cost breakdown shown in \Tab{supp:computational_costs} reveals that computing the uncertainty estimates for \code{LTAU-FF} occupies about 1\% of the walltime for \code{3BPA} and \code{Carbon\_GAP\_20}, or 7\% for \code{OC20}.
This can be compared to the 15-50\% for graph construction and 40-80\% for the model forward pass.

While \code{LTAU-FF} is 10-50$\times$ cheaper than the cost of a forward pass of the model, \code{Ensemble} can be expected to be approximately $M\times$ the cost of the forward pass.
Given that ensemble sizes of $5 \leq M \leq 30$ have been observed to be necessary for \code{Ensemble} \cite{Gastegger2017,Lu2023}, this means that \code{LTAU-FF} can be expected to be 2--3 orders of magnitude faster than \code{Ensemble} in practice for inference.
Furthermore, since \code{LTAU-FF} only requires the training trajectory from a single model, it can be easily applied to tasks like the \code{OC20} dataset where it is impractical to train multiple models.
Considering that many practical applications of atomistic force fields involve simulating systems with millions or even billions of atoms, and that so-called ``universal'' interatomic potentials \cite{Chen2022,Batatia2024} require training to massive datasets, we believe that \code{LTAU-FF} will be an essential tool for UQ in this field.

\section{Conclusion}

In this work, we developed \code{LTAU}, a novel UQ method based on the per-sample PDFs of training errors and a distance-based similarity search, and demonstrated its utility in the field of atomistic simulations (\code{LTAU-FF}).
We outlined how \code{LTAU-FF}, which avoids many of the drawbacks of previous ensemble-based UQ methods for atomistic force fields, can be readily applied to any model while introducing only negligible computational overhead and requiring only minor modifications to training software.
Finally, we show that the UQ metric produced by \code{LTAU-FF}
outperforms the gold standard ensemble method on ID data, has promising avenues for detecting OOD data,
% outperforms the gold standard ensemble method on almost all test,
% can be calibrated using a simple scaling factor,
and can be used as a reliable tool for re-weighting training sets and predicting model failure during simulations.

Further research coupling \code{LTAU} with more advanced clustering methods, error propagation techniques, or other modifications that may improve the diversity of the ensemble even more (e.g., alternative loss functions or training regimen) would be particularly welcome contributions.
We are also interested in exploring methods beyond those proposed in \Sec{calibration} for improving the ability of \code{LTAU-FF} to detect OOD data and make more accurate predictions on near-OOD data.
% applying \code{LTAU} to other deep learning regressio tasks
Finally, we point out that although in this work we only assessed \code{LTAU} as applied to atomistic force fields, the mathematical foundations of our method allow it to be applied to any regression task, and we fully intend to explore its use in other machine learning applications in future work.
% Moving forward, we hope that this work will serve as a valuable tool for developing machine-learned atomistic force fields and other machine learning models.
% by aiding in the design of active learning workflows, failure analysis methods, and data pruning techniques.

\section{Data and code availability}
\label{sec:availability}
All training sets can be found at their original sources.
The \code{LTAU-FF} code and all models developed in this work will be provided upon completion of the peer review of this paper. Code for computing miscalibration areas is available at at \url{https://github.com/uncertainty-toolbox/uncertainty-toolbox/} under the MIT license. Code for training the NequIP models is available at \url{https://github.com/mir-group/nequip} under the MIT license.

\section{Author contributions}
\textbf{Joshua A. Vita}: Conceptualization, Methodology, Software, Validation,  Formal analysis, Investigation, Data Curation, Writing - Original Draft, Writing - Review \& Editing, Visualization.
\textbf{Amit Samanta}: Methodology, Writing - Review \& Editing, Supervision
\textbf{Fei Zhou}: Methodology, Writing - Review \& Editing, Supervision
\textbf{Vincenzo Lordi}: Methodology, Resources, Writing - Review \& Editing, Supervision, Project administration, Funding acquisition.

\section{Acknowledgments}
This work was performed under the auspices of the U.S. Department of Energy by Lawrence Livermore National Laboratory under Contract DE-AC52-07NA27344, funded by the Laboratory Directed Research and Development Program at LLNL under project tracking code 23-SI-006.
The authors would also like to thank Jared Stimac for providing useful discussions on comparing UQ metrics.

\bibliographystyle{unsrtnat}  % longer in-line stuff;; no it's not? this gives in-order numbers
\bibliography{bibliography}

\appendix
\onecolumn

\renewcommand{\thefigure}{A\arabic{figure}}
\setcounter{figure}{0}
\renewcommand{\thetable}{A\arabic{table}}
\setcounter{table}{0}

\section{Additional details on computational cost}

All experiments in this work were performed on a single node of the Sierra supercomputer. Each node has 256 GB of CPU memory, 44 CPU cores, and four NVIDIA V100 GPUs (though only one was used due to a lack of multi-GPU support in the NequIP code). Training a single NequIP model on this architecture required walltimes of approximately 4 hours for the \code{3BPA} dataset and 48 hours for the \code{Carbon\_GAP\_20} dataset. Training to the \code{200k} split of the \code{S2EF} task from \code{OC20}, which is a significantly larger dataset, required approximately 500 hours.

\begin{table*}[ht!]
\centering
\caption{Computational costs of evaluating model uncertainty for the datasets used in this work. Results were obtained by computing forces and uncertainties for a single NequIP model on a random selection of 500 test samples for each dataset. \code{LTAU-FF} used $k=10$ neighbors and descriptors of dimensionality 32 for all datasets.}
\label{tab:supp:computational_costs}
\begin{tabular}{cclr}
\hline \hline
Dataset & Train size (\# atoms) & Operation & \% time \\
\hline
 & & UQ (\code{IndexFlatL2}) & 1.5 \\
 \code{3BPA} & 13,500 & Graph construction & 47.0 \\
 & & Forward pass & 51.4 \\
 \hline
%   & & UQ (\code{IndexHNSWFlat}, $n=10$) & 0.4 \\
%  \code{Carbon\_GAP\_20} & 400,275 & Graph construction & 35.1 \\
% & & Forward pass & 64.4 \\
%   & & UQ (\code{IndexHNSWFlat}, $n=10$) & 3.4 \\
%  \code{Carbon\_GAP\_20} & 400,275 & Graph construction & 33.9 \\
% & & Forward pass & 62.2 \\
& & UQ (\code{IndexHNSWFlat}) & 1.0 \\
 \code{Carbon\_GAP\_20} & 400,275 & Graph construction & 53.3 \\
& & Forward pass & 45.4 \\
 \hline
  & & UQ (\code{IndexHNSWFlat}) & 7.3 \\
 \code{OC20} & 14,631,937 & Graph construction & 14.2 \\
& & Forward pass & 78.0 \\
\hline \hline
\end{tabular} \end{table*}

\renewcommand{\thefigure}{B\arabic{figure}}
\setcounter{figure}{0}
\renewcommand{\thetable}{B\arabic{table}}
\setcounter{table}{0}

\section{Hyperparameter details}
\label{sec:supp:hyperparams}

\subsection{NequIP}

\Tab{hyper_params} outlines some of the most important parameters used for defining the NequIP architecture, the training algorithm, or computing the uncertainty estimates for each dataset.
Unless otherwise specified, the recommended settings were used for the NequIP model as provided by \url{https://github.com/mir-group/nequip/blob/main/configs/full.yaml}.
% For \code{OC20} the 200K split of the training data for the \code{S2EF} task was used with no validation set.
The \code{AMSGrad} variant of the \code{Adam} optimizer was used with an initial learning rate of 0.005, a weight decay of 0, and the \code{ReduceLROnPlateau} scheduler.
Full configuration files will be provided upon completion of the peer review of this article.

\begin{table*}[ht!]
\centering
\caption{The main hyperparameters used for the NequIP model. \code{r\_max} is the radial cutoff of the model, \code{num\_layers} is the number of message passing layers, \code{l\_max} is the symmetry order of the equivariant features, and $M$ is the ensemble size.}
\label{tab:hyper_params}
\begin{tabular}{cccccccc}
\hline \hline
Dataset & \code{r\_max} & \code{num\_layers} & \code{l\_max} & F weight & E weight & Batch size & $M$ \\
\hline
 \code{3BPA} & 5.0 & 5 & 3 & 1000.0 & 0.0 & 5 & 10\\
 \code{Carbon\_GAP\_20} & 4.5 & 4 & 2 & 10.0 & 0.0 & 1 & 10 \\
 \code{OC20} & 4.0 & 2 & 2 & 100.0 & 1.0 & 4 & 1\\
\hline \hline
\end{tabular} \end{table*}

\subsection{FAISS}

The cost of UQ predictions with \code{LTAU-FF} is dictated by the complexity of the similarity search performed by the \code{FAISS} package, as well as the size of the training set and the dimensionality of the latent space embeddings used as the atomic descriptors.
Depending on these factors, particularly the dataset size, different indexing algorithms (as implemented by \code{FAISS}) are recommended.
For the \code{Carbon\_GAP\_20} and \code{OC20} datasets, the approximate neighbor search \code{IndexHNSWFlat} index was used as opposed to the exact brute-force approach (\code{IndexFlatL2}) used for \code{3BPA}.
Additional UQ speedups could be obtained by using the GPU implementation of FAISS, decreasing $k$, or further refining the parameters of the indexers.
For a more thorough discussion of indexing methods and hyperparameter choices, we recommend consulting the \code{FAISS} documentation.

While the \code{IndexFlatL2} index is a parameter-free brute force approach, the \code{IndexHNSWFlat} index has two primary parameters which strongly affect the cost of the similarity search.
The Heirarchical Navigable Small Worlds (HNSW) method \cite{Malkov2016} is an \textit{approximate} similarity search algorithm which decomposes the search space into a multi-layered graph structure.
The key parameters of \code{IndexHNSWFlat} are \code{M} (the number of neighbor links to add for each point in the graph), \code{efConstruction} (the number of neighbors to consider at each layer when inserting into the graph), and \code{efSearch} (the number of neighbors to consider at each layer when searching the graph).
In this work, we used values of \code{M}$=32$, \code{efConstruction}$=40$, and \code{efSearch}$=16$, which we found to provide a reasonable balance between speed and accuracy.
The computational cost of model evaluation using these hyperparameters are shown in \Tab{supp:computational_costs}.
We emphasize that more thorough analysis of the effects of tuning these hyperparameters would be helpful.

\renewcommand{\thefigure}{C\arabic{figure}}
\setcounter{figure}{0}
\renewcommand{\thetable}{C\arabic{table}}
\setcounter{table}{0}

\section{Dataset details}
\label{sec:supp:data}

\subsection*{\code{3BPA}}
The first dataset used in this work is the \code{3BPA} dataset \cite{Kovcs2021}, which has been used previously in the literature for benchmarking extrapolation behaviors of force fields \cite{Batatia2022a,Batatia2022b,Musaelian2023}.
The \code{3BPA} dataset is a molecular dataset consisting of 500 training configurations taken from molecular dynamics (MD) simulations at a temperature of 300 K, where each configuration has 27 atoms for a total of 13,500 training points.
There are additionally three separate test sets sampled using MD simulations at temperatures of 300K, 600K, and 1200K, respectively, where increasing the sampling temperature can be expected to push the simulations to explore OOD regions of the available configuration space of the molecule.
For \code{3BPA} the predefined training set was used with no validation set.
The dataset can be obtained from \url{https://pubs.acs.org/doi/10.1021/acs.jctc.1c00647} under the CC BY 4.0 license.

\subsection*{\code{Carbon\_GAP\_20}}
In order to better understand how \code{LTAU-FF} performs on a more challenging dataset, we also used the \code{Carbon\_GAP\_20} dataset \cite{Rowe2020}, which was originally intended to be used for fitting a model capable of accurately describing a broad range of carbon phases.
The \code{Carbon\_GAP\_20} dataset includes a training set of 6,088 configurations (400,275 atoms), as well as a larger superset containing 17,525 configurations (1,345,246 atoms) which we used for testing.
The training and test sets are comprised of an extremely diverse range of phases, including bulk crystals, amorphous carbon, graphene, graphite, fullerene, nanotubes, liquids, surfaces, defected configurations, and other rare allotropes obtained through random structure search or extracted from the literature.
Particularly valuable are the labels provided by the authors of the \code{Carbon\_GAP\_20} dataset for defining conceptually similar clusters over the atomic configurations in the dataset.
Although in our analysis we observed a number of mislabeled configurations, the provided groupings are still extremely useful for understanding how models and methods perform across different subsets of the data.
A breakdown of the number of atoms in each group in the training set is provided in \Tab{supp:config_types}.
A more thorough description of the dataset can be found in \cite{Rowe2020}.
For \code{Carbon\_GAP\_20} a random 90--10 training--validation split was used.
The dataset can be obtained from \url{https://www.repository.cam.ac.uk/handle/1810/307452} under the CC BY 4.0 license.

\begin{table*}
\centering
\caption{Number of atomic configurations and atoms within each group of the \code{Carbon\_GAP\_20} training set.}
\label{tab:supp:config_types}
\begin{tabular}{lrr}
\hline \hline
\code{config\_type} group & \# configurations & \# atoms \\
\hline
 \code{Amorphous\_Bulk} & 3,053 & 200,470 \\
 \code{Amorphous\_Surfaces} & 20 & 2,648 \\
 \code{Crystalline\_Bulk} & 78 & 368 \\
 \code{Crystalline\_RSS} & 483 & 8,554 \\
 \code{Defects} & 530 & 70,022 \\
 \code{Diamond} & 164 & 1,360 \\
 \code{Dimer} & 26 & 52 \\
 \code{Fullerenes} & 272 & 11,782 \\
 \code{Graphene} & 2 & 400 \\
 \code{Graphite} & 185 & 6,344 \\
 \code{Graphite\_Layer\_Sep} & 7 & 700 \\
 \code{LD\_iter1} & 80 & 14,048 \\
 \code{Liquid} & 6 & 1,296 \\
 \code{Liquid\_Interface} & 17 & 3,672 \\
 \code{Nanotubes} & 138 & 4,976 \\
 \code{SACADA} & 755 & 24,503 \\
 \code{Surfaces} & 271 & 49,079 \\[5pt]
 \code{Total} & 6,088 & 400,275 \\
\hline \hline
\end{tabular} \end{table*}

\subsection*{\code{OC20}}
As a practical application, we also train a model to the \code{S2EF} task of the \code{OC20} dataset \cite{Chanussot2021} then test how well the UQ estimate provided by \code{LTAU-FF} predicted performance on the \code{ISR2S} task.
Specifically, we trained to the \code{200k} split (14,631,937 atoms) of the \code{S2EF} task, and tested on the \code{val\_id} and \code{val\_ood\_both} splits of the \code{IS2RS} task.
The \code{OC20} dataset includes a wide range of catalysts comprised of various materials, surfaces, and adsorbates.
Importantly, the data includes labels for each atom identifying them as an ``adsorbate'' atom (part of the adsorbing molecule), a ``surface'' atom (the top few layers of the material), or a ``bulk'' atom (everything in the material that is not part of the surface).
For \code{OC20} the predefined \code{S2EF} training set and \code{IS2RS} validation sets were used.
The dataset can be obtained from \url{https://fair-chem.github.io/core/datasets/oc20.html} under the MIT license.

\clearpage
\renewcommand{\thefigure}{D\arabic{figure}}
\setcounter{figure}{0}
\renewcommand{\thetable}{D\arabic{table}}
\setcounter{table}{0}

% \section{Additional analysis of uncertainty}
% \label{sec:supp:add_analysis}

\section{Out-of-domain detection}
\label{sec:supp:ood}

\begin{figure}[ht!]
\centering
\includegraphics[width=\linewidth]{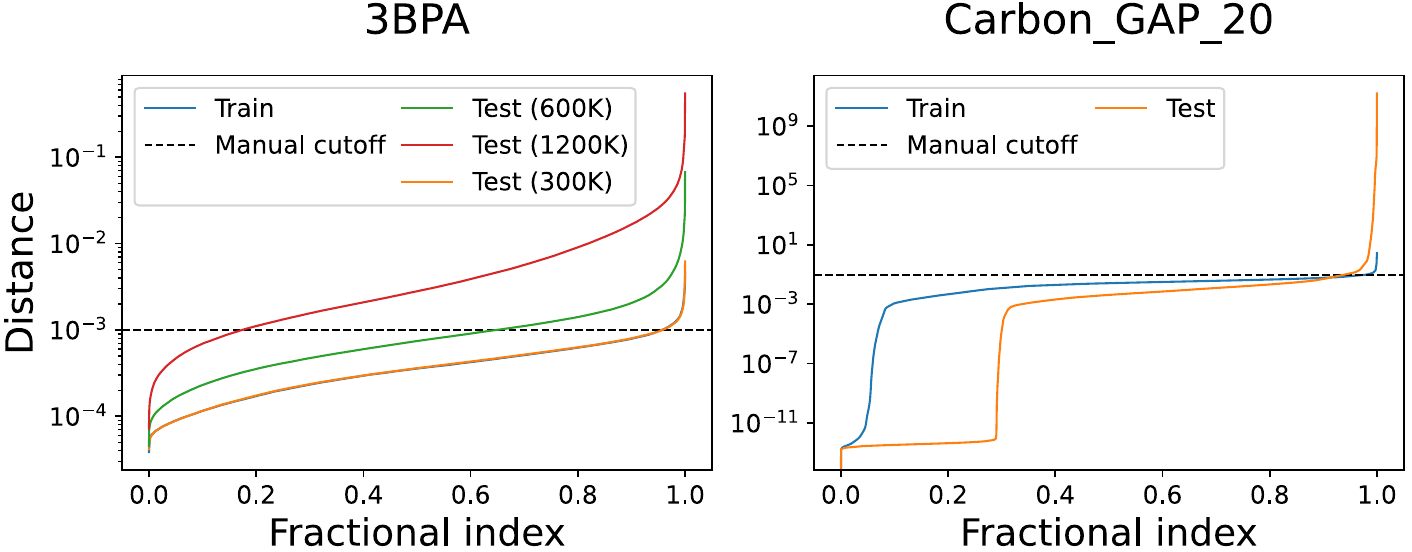}
\caption{
Distributions of first nearest-neighbor distances in the train/test sets of \code{3BPA} and \code{Carbon\_GAP\_20}.
The data points are sorted by distance to improve visibility.
Each train/test set is plotted as a different color; the \code{3BPA} 300K test data almost exact overlaps the \code{3BPA} train data.
Cutoffs (dashed black lines) can be chosen \textit{ad hoc} to identify OOD data where \code{LTAU-FF} may begin to suffer from the drawbacks of distance-based UQ metrics discussed in \Sec{distance_uq}.
\Fig{supp:3bpa_ood} and \Fig{supp:gap20_ood} show that data classified as being OOD roughly corresponds to the points where \code{LTAU-FF} fails to accurately predict the true force errors.}
\label{fig:supp:distances}
\end{figure}

\begin{figure*}[ht!]
\centering
\includegraphics[width=\linewidth]{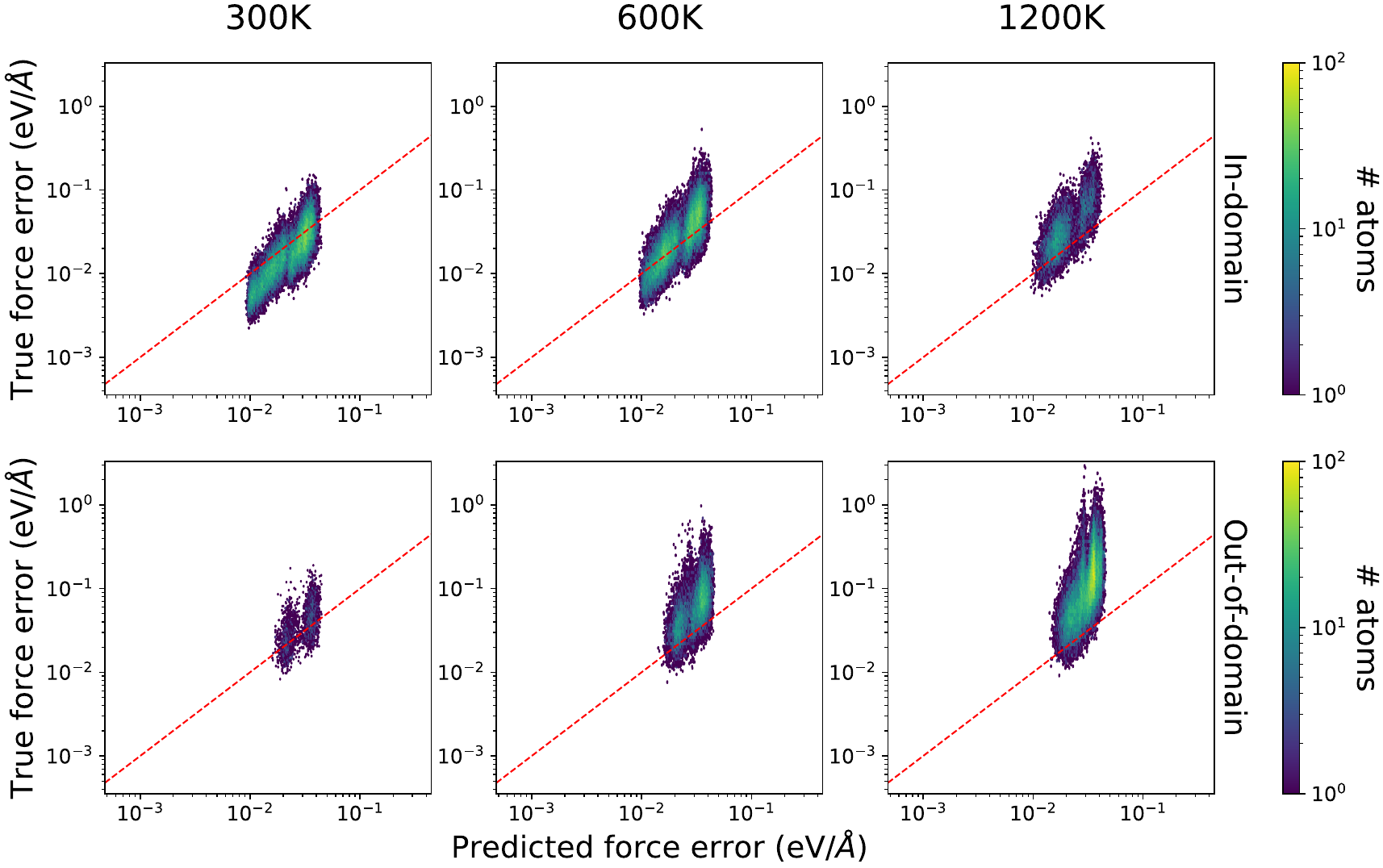}
\caption{A version of \Fig{parity_cal_3bpa}a where the \code{3BPA} test data is split into ID and OOD data based on the manual cutoffs defined in \Fig{supp:distances}.}
\label{fig:supp:3bpa_ood}
\end{figure*}

\begin{figure*}[ht!]
\centering
\includegraphics[width=0.75\linewidth]{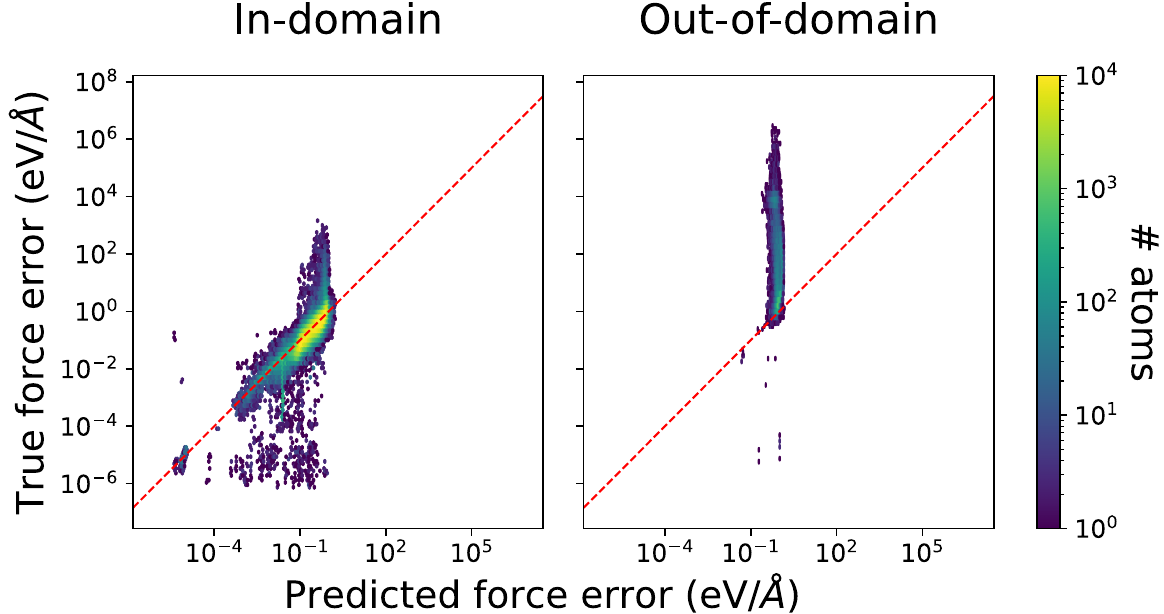}
\caption{A version of \Fig{parity_cal_gap20}a where the \code{Carbon\_GAP\_20} test data is split into ID and OOD data based on the manual cutoffs defined in \Fig{supp:distances}.}
\label{fig:supp:gap20_ood}
\end{figure*}

\renewcommand{\thefigure}{E\arabic{figure}}
\setcounter{figure}{0}
\renewcommand{\thetable}{E\arabic{table}}
\setcounter{table}{0}

\clearpage
\section{Dataset re-weighting}
\label{sec:supp:upweight}

% In order to re-weight the \code{Carbon\_GAP\_20} training data, we chose to define the ``difficulty'' of a sample in a manner similar to the sample ``confidence'' used elsewhere in the literature for classification tasks \cite{Swayamdipta2020}.
In \cite{Swayamdipta2020}, a model is said to be ``confident'' in its prediction for a given point if the point has a high probability of being correctly predicted (i.e., high average Softmax activation) throughout the course of training.
Since regression tasks do not normally have a Softmax output layer and there is no binary notion of ``correctness'', we instead chose to define the ``difficulty'' of a sample as the likelihood (over the course of training) of it having errors above the final mean absolute training error (MAE).
Since \code{LTAU-FF} already requires recording the PDFs of per-sample errors, this difficulty metric can be easily computed.

\begin{figure}[ht!]
\centering
\includegraphics[width=0.5\linewidth]{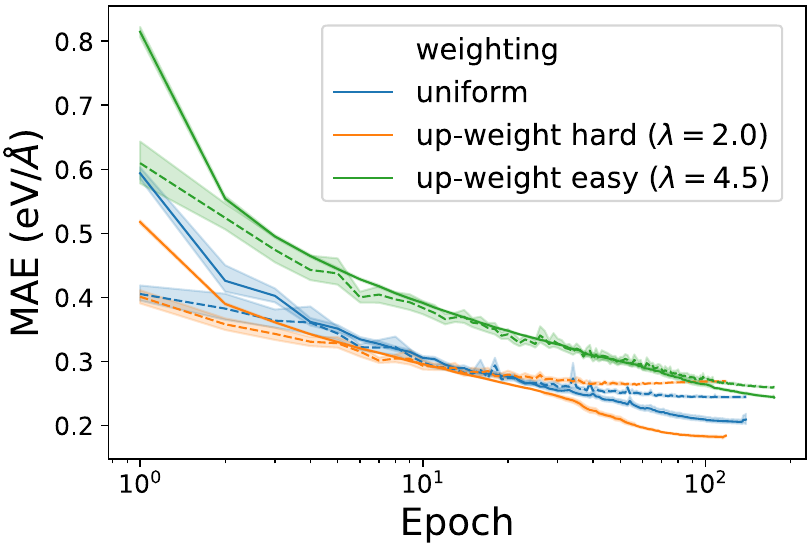}
\caption{Training (solid lines) and validation (dashed lines) curves for ensembles of 10 models trained to the \code{Carbon\_GAP\_20} dataset with different weighting schemes. A random 90:10 training:validation split was used, where all runs used the exact same split. Shaded bands denote the 95\% confidence intervals computed across the 10 runs for each weighting scheme.
% Increasing the weight on samples (as described in \App{supp:upweight}) with low $\mu_M(p_i)$ values (``up-weight hard'') increased the degree of overfitting, while increasing the weight on samples with high $\mu_M(p_i)$ (``up-weight easy'') decreased overfitting.
% A scaling factor of $\lambda=2.0$ for re-weighting the samples (as described in \App{supp:upweight}) was used initially for the ``up-weight easy'' runs, but was later increased to 4.5 because the lower value of 2.5 resulted in runs which were nearly indistinguishable from the ``uniform'' weighting scheme.
% The ``up-weight easy'' models were trained for about 25\% longer to improve convergence of the training--validation gap; even longer runs may have widened the gap further, but would have required an unreasonably high computational budget.
}
\label{fig:supp:training_curves}
\end{figure}

In the case of the ``up-weight hard'' weighting scheme in \Fig{supp:training_curves}, atom $i$ was given a weight in the loss function of $w_i = \exp [\lambda (1 - d_i)]$, where $d_i \coloneqq P(\epsilon_i < \text{MAE})$.
In the ``up-weight easy'' scheme, the weight was given as $w_i = \exp (\lambda d_i)$.
Values of $\lambda$ were determined \textit{ad hoc}, and set to 2.0 and 4.5 for the up-weight easy and hard schemes, respectively.
% \Fig{supp:upweight} provides a visual depiction of the two weighting schemes.
% Notably, the relationship between $\mu_M(p_i)$ and $\sigma_M(p_i)$ observed in \Fig{supp:upweight} exactly follows the expected pattern from \cite{Swayamdipta2020}, where the dataset can be roughly divided into ``easy-to-learn'', ``hard-to-learn'', and ``ambiguous'' subsets.

% \begin{figure*}[ht!]
% \centering
% \includegraphics[width=\linewidth]{figures/supp/upweight.pdf}
% \caption{Visualization of the re-weighting schemes used for the \code{Carbon\_GAP\_20} dataset in \Fig{supp:training_curves}. Scaling factors, $\lambda$, were selected \textit{ad hoc} based on experimentation in order to obtain weights which resulted in a noticeable change in the training--validation gap as shown in \Fig{supp:training_curves}.}
% \label{fig:supp:upweight}
% \end{figure*}

\begin{figure*}[ht!]
\centering
\includegraphics[width=\linewidth]{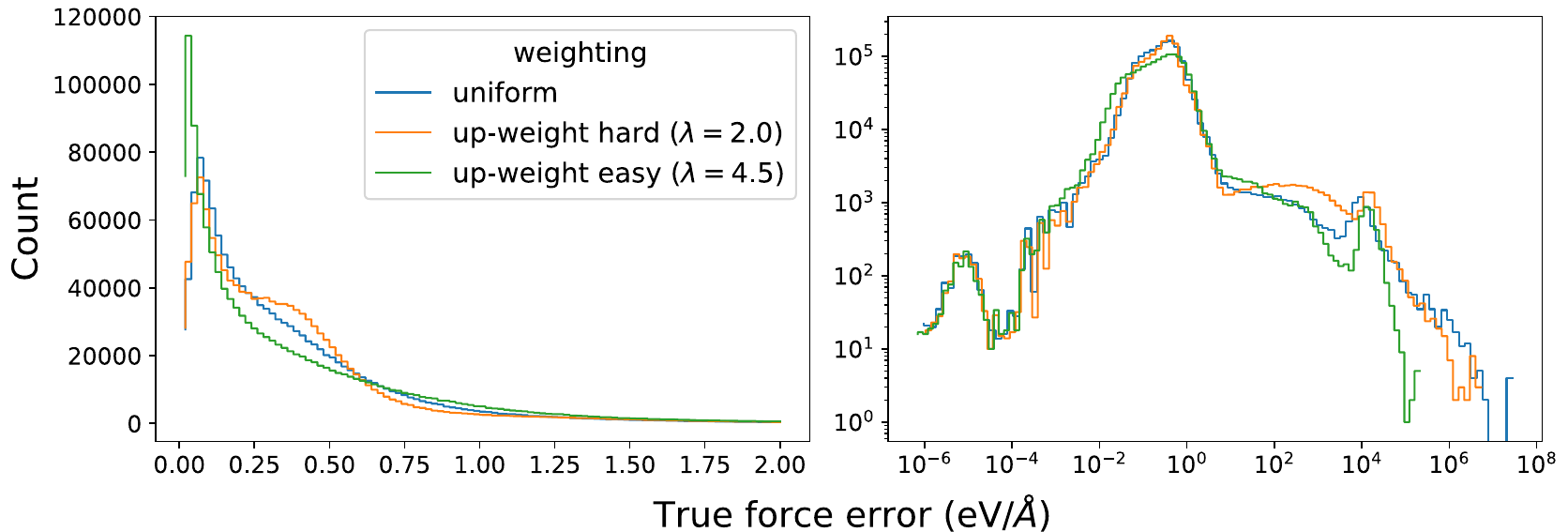}
\caption{Distributions of test errors on the \code{Carbon\_GAP\_20} dataset for the three weighting schemes. In panel \textbf{a}, the range of the x-axis has been clipped to a maximum value of 2.0 eV/\AA~ in order to improve visualization by removing a small number (3-4\%) of points with abnormally large errors. In panel \textbf{b}, the full distributions are shown on a log-scale. The average test error for all weighting schemes after removing values larger than 2.0 eV/\AA, was around 0.35 eV/\AA. However, when including all data from panel \textbf{b}, the ``up-weight easy'' scheme had an MAE of 47eV/\AA~, which was an order of magnitude lower than the other two schemes.}
\label{fig:supp:upweight_errors}
\end{figure*}

\end{document}